\documentclass[10pt,twocolumn,letterpaper]{article}

\usepackage{cvpr}
\usepackage{times}
\usepackage{epsfig}
\usepackage{graphicx}
\usepackage{amsmath}
\usepackage{amssymb}
\usepackage{multirow}
\usepackage[usenames,dvipsnames]{color}
\usepackage{caption}
\usepackage{soul}
\usepackage{fixltx2e}
\usepackage{subcaption}

\usepackage[pagebackref=true,breaklinks=true,letterpaper=true,colorlinks,bookmarks=false]{hyperref}

 \cvprfinalcopy 

\ifcvprfinal\fi
\begin{document}

\title{ Preserving Semantic Relations for Zero-Shot Learning}

\author{Yashas Annadani\thanks{Currently at IIIT-H. Majority of the work was completed when Yashas was at NITK and was an intern at IISc.}\\
	National Institute of Technology - Karnataka\\
	{\tt\small yashas.annadani@gmail.com}
	\and
	Soma Biswas\\
	Indian Institute of Science\\
	{\tt\small somabiswas@iisc.ac.in}
}

\maketitle


\begin{abstract}
Zero-shot learning has gained popularity due to its potential to scale recognition models without requiring additional training data. 
This is usually achieved by associating categories with their semantic information like attributes. However, we believe that the potential offered by this paradigm is not yet fully exploited. 
In this work, we propose to utilize the structure of the space spanned by the attributes using a set of relations.
We devise objective functions to preserve these relations in the embedding space, thereby inducing semanticity to the embedding space. 
Through extensive experimental evaluation on five benchmark datasets, we demonstrate that inducing semanticity to the embedding space is beneficial for zero-shot learning. 
The proposed approach outperforms the state-of-the-art on the standard zero-shot setting as well as the more realistic generalized zero-shot setting.
We also demonstrate how the proposed approach can be useful for making approximate semantic inferences about an image belonging to a category for which attribute information is not available.

\end{abstract}

\section{Introduction}

Novel categories of objects arise dynamically in nature. It is estimated that around 8000 species of animals and plants are discovered every year~\cite{numObj}.  
However, current recognition models are quite incapable of handling this dynamic scenario when labeled examples of novel categories  are not available. 
Obtaining labeled examples followed by retraining or transfer learning can be expensive and cumbersome. 
Zero-shot learning  (ZSL)~\cite{palatucci2009zero,akata2013label,norouzi2013zero,socher2013zero,zhang2016zero,lampert2014attribute} offers an elegant way to address this problem by utilizing the mid-level semantic descriptions of the categories.  These descriptions are usually encoded in an \textit{attribute} vector~\cite{ferrari2008learning,farhadi2009describing,farhadi2010attribute}, sometimes referred to as side information or class embeddings. 
This paradigm is useful not only for emerging categories, but also for extending the recognition capabilities of a model beyond the categories it is trained on without requiring additional training data.
\begin{figure}[t]
	\centering
	\includegraphics[ height=5.5cm]{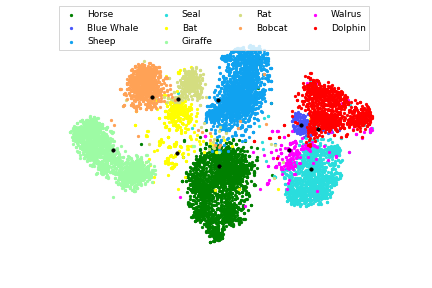}
	\caption{\small t-SNE visualization \cite{maaten2008visualizing} of the ten unseen categories of AWA2 \cite{xian2017zero1} dataset. Black dot denotes the mapped embeddings obtained using the proposed approach. Best viewed in color.}
	\label{tsne-awa} 
\end{figure} 

Some of the existing approaches treat zero-shot recognition as a ranking problem~\cite{akata2016label,xian2016latent,frome2013devise}. In these approaches, a compatibility function between the image feature and the class embeddings is learned such that its score for the correct class is higher than that for an incorrect class by a fixed margin. 
Ranking can lead to loss of some of the semantic structure available from the attributes due to the fixed margin and the unbounded nature of the compatibility function. 
In some of the other approaches~\cite{Kodirov_2017_CVPR,zhang2016learning,romera2015embarrassingly}, zero-shot recognition is typically achieved by embedding either the image features or the attribute vectors, or both, to a predefined embedding space using a ridge regression or a mean squared error objective.
Here, proper choice of the embedding space is essential.
If the space spanned by the attributes (semantic space) is used as the embedding space, then the semantic  structure is preserved, but the problem of \textit{hubness} surfaces~\cite{shigeto2015ridge,radovanovic2010hubs}.
To alleviate this problem, few recent approaches ~\cite{zhang2016learning, shigeto2015ridge} map the class embeddings to  the space spanned by the image features (visual space). 
However, the visual space may not contain semantic properties, as it may be inherited from a model trained on a supervised classification task where labels are one-of-k coded.

We believe that two things are crucial for zero-shot recognition: (1) \textit{discriminative ability} on the categories available during training and (2) \textit{inheriting the properties} of semantic space for efficient classification on novel categories.  
Existing approaches focus on either of the two aspects.
To this end, we propose a simple yet effective approach which ensures discriminative capability while retaining the structure of the semantic space in an encoder-decoder multilayer perceptron framework. 

We decompose the structure of the semantic space to a set of relations between categories. 
Our aim is to preserve these relations in the embedding space so as to appropriately inherit the structure of the semantic space to the embedding space. 
Relation between categories is decomposed to three groups: identical, semantically similar and semantically dissimilar.
We construct a semantic tuple which consists of samples belonging to categories of each of the relations with respect to a given category. 
Objective function specific to each relation is formulated so that the underlying semantic structure can be captured while still ensuring discriminative capability. 
The underlying principle is that the  embeddings belonging to categories which are semantically similar in the attribute space must still be close in the embedding space, while the ones which are dissimilar should be far away. 

Our contributions are threefold:
\begin{itemize}
	\itemsep0em
	\item Propose a simple and effective approach for zero-shot recognition which preserves the structure of the semantic space in the embedding space by utilizing semantic relations between categories.
	\item Extensive experimental evaluation on multiple datasets, including the large scale ImageNet~\cite{deng2009imagenet} shows that the proposed method improves over the state-of-the art in multiple settings.
	\item Demonstrate how the proposed approach can be useful for making approximate inferences about images belonging to novel categories even when the class embeddings corresponding to that category is not available. 
\end{itemize}
The rest of the paper is organized is follows: Section~\ref{related} reviews the related work followed by Section~\ref{proposed} which describes the proposed approach in detail. Experimental evaluation is reported in Section~\ref{expts} with some pertinent discussions in Section~\ref{discuss}. 
We conclude the paper in Section~\ref{conclude}.

\section{Related Work}
\label{related}
There are a number of related works which have been developed independently to address zero-shot recognition for visual data. Few of these approaches use bilinear compatibility frameworks 
to model the relationship between the features and the class embeddings. Akata \etal~\cite{akata2016label,akata2013label} and Frome \etal \cite{frome2013devise} use a pairwise ranking formulation to learn the parameters of the bilinear model. 
~\cite{akata2015evaluation} use distributed word embeddings like Word2Vec~\cite{mikolov2013distributed} and Glove~\cite{pennington2014glove} apart from annotated attributes to learn a bilinear model for each of them independently. The final model is obtained by a weighted combination of the individual compatibility frameworks. Xian \etal \cite{xian2016latent} learn multiple bilinear models that results in a piece-wise linear decision boundary, which suits fine-grained classification. Romera-Pardes \etal~\cite{romera2015embarrassingly} map the resulting compatibility model to label space. This method is simple and elegant as it has a closed form solution. Qiao \etal \cite{qiao2016less} extend this method to online documents by incorporating an $\ell_{2,1}$ norm on the weight vector to suppress noise in the documents and enhance zero-shot recognition. Although bilinear compatibility models are elegant to use, the bilinear compatibility score which is obtained at the time of inference has limited semantic meaning, which restricts its interpretability. 

Apart from bilinear compatibility models, few other approaches map image features to semantic space by using a ridge regression objective. 
Kodirov \etal \cite{Kodirov_2017_CVPR} use an additional reconstruction constraint on the mapped features which enhanced zero-shot recognition performance. In \cite{lei2015predicting}, the parameters of a deep network is learned using side information like Word2Vec. This work uses binary cross entropy loss and hinge loss in addition to mean squared error loss. Recently, Zhang \etal\cite{zhang2016learning} proposed to reverse the direction of mapping from semantic space to visual space. 
However, mapping from semantic space to visual space may result in reduced semantic expressiveness of the model.

Few of the other approaches employ manifold learning \cite{changpinyo2016synthesized,Xu_2017_CVPR, Morgado_2017_CVPR} to solve zero-shot recognition. Changpinyo \etal \cite{changpinyo2016synthesized} construct a weighted bipartite graph in a space where additional classes called phantom classes are introduced. They minimize a distortion objective which aligns this space with the class embeddings. \cite{Xu_2017_CVPR} use a transductive approach to learn projection function by matrix tri-factorization and preserving the underlying manifold structure of both visual space and semantic space.
However, our approach differs from these approaches as our method is neither transductive nor involves manifold learning.

Our method relies on using semantic relations to learn the embeddings. Parikh and Grauman \cite{parikh2011relative} use partial ordering and ranking formulation to capture semantic relations across individual attributes. This is different from our approach wherein the semantic relations are defined on the categories themselves.
\begin{figure*}[t!]
\centering
\includegraphics[width=0.9\textwidth,height=5.7cm]{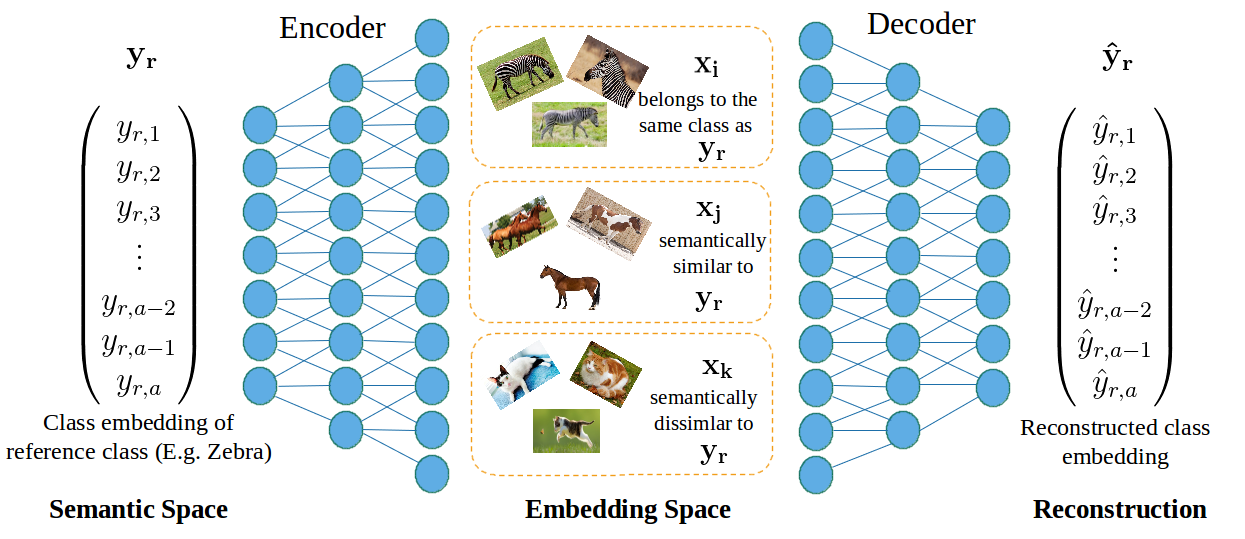}
\caption{\small An illustration of the proposed approach.}
\label{main}
\end{figure*}


\section{Proposed Approach}
\label{proposed}
Let $\mathbf{x_i}$ be a $d$-dimensional feature descriptor of an image sample corresponding to one of the seen classes $C_s =\{1,2,\dots,c_s\}$. The set of all samples in the given training data is denoted by $\mathbf{X} = \{\mathbf{x_i}^T\}_{i=1}^{N}$$\in \mathbb{R}^{N\times d}$ and their corresponding class embeddings by $\mathbf{Y} =\{\mathbf{y_i}^T\}_{i=1}^{N}$$\in \mathbb{R}^{N\times a}$, where $a$ is the dimensionality of the class embedding. In this work, the class embeddings are obtained either through attributes~\cite{farhadi2009describing,lampert2009learning,wah2011caltech,xiao2010sun} or distributed word representations~\cite{mikolov2013distributed}. 
The semantic space spanned by the class embeddings is shared between seen and unseen classes.  
Given a new sample $\mathbf{x}^u$ which potentially belongs to an unseen class $C_u=\{c_s+1,\dots,c_s+c_u\}$,  the goal of zero-shot learning is to predict the correct class of $\mathbf{x}^u$, without ever having trained the recognition model using the samples of the unseen class. 
\subsection{Defining Semantic Relations}
\label{relations}
We explicitly define semantic relations between classes so that objective function specific to each relation can be formulated in order to preserve the relations. For a given set of classes, we wish to group them to three different categories with respect to a reference class - \textit{identical}, \textit{semantically similar} and \textit{semantically dissimilar}. This particular grouping should be reflective of the underlying semanticity. There are many possible ways to define these class relations. For example, one can leverage prior knowledge about the classes specific to the task under study. 

In this work, we employ class embeddings to 
define semantic relations. Let $\delta_{mn}=s(\mathbf{y_m},\mathbf{y_n})$ be a similarity measure between two class embeddings. We use cosine similarity as a semantic similarity measure~\cite{mikolov2013distributed,mikolov2013efficient}.
\begin{equation}
s(\mathbf{p},\mathbf{q}) = \dfrac{<\mathbf{p}\, ,\, \mathbf{q}>}{||\mathbf{p}||_2||\, \mathbf{q}||_2}
\end{equation}
$\mathbf{p}$ and $\mathbf{q}$ are any two vectors of same dimension.
Given $\delta_{mn}$ for any two embeddings $\mathbf{y_m}$ and $\mathbf{y_n}$, we define class relations as follows:
\begin{itemize}
	\itemsep0em
	\item Belonging to same class (\textit{identical}) if $\delta_{mn}  =1$.
	\item \textit{Semantically similar} if $ \tau \leq \delta_{mn} < 1$.
	\item \textit{Semantically dissimilar} if $\delta_{mn}$ $<$ $\tau$.
\end{itemize}
where $\tau\in (-1,1)$ is a threshold. Without loss of generality, $\tau$ can be fixed to zero, which is a reasonable estimate for a cosine similarity. $\tau$ can also be chosen over a validation set. 
In this paper, we choose $\tau$ based on the performance on the validation set.
Since the semantic space of attributes is shared between seen and unseen classes, this particular definition provides a good generalization to novel classes.
\subsection{Preserving Semantic Relations}
Armed with the above definition, we wish to map the class embeddings to the visual space such that the semantic
relation between the mapped class embeddings and the visual features reflects the relation between their corresponding classes. Our motivation to map to the visual space comes from the works of Shigeto \etal~\cite{shigeto2015ridge}\footnote{\cite{shigeto2015ridge} discusses hubness problem only for ridge regression based techniques. However, hubness problem can also arise in cosine similarity measure based approaches, as demonstrated in \cite{radovanovic2010existence}.} 
and Zhang \etal~\cite{zhang2016learning}, which showed that using visual space instead of semantic space or any other intermediate space as the embedding space alleviated the \textit{hubness} problem~\cite{tomasev2014role,radovanovic2010hubs}, a problem in which a few points (called as hubs) arise which are in the $k$-nearest neighbors of most of the other points. We employ nearest neighbor search for zero-shot recognition, and hence the problem of hubness can lower the performance if mapped to any space other than the visual space. Therefore, we use the visual space as the embedding space for our approach. 

We use an encoder-decoder multilayer perceptron architecture to learn the embedding function. The encoder parameterized by $\theta_f$ learns the mapping $f(\mathbf{y};\theta_f)$ and the decoder $g(\mathbf{x};\theta_g)$ learns to reconstruct the input from the mapped class embeddings. Our model formulation is inspired from~\cite{Kodirov_2017_CVPR}, though~\cite{Kodirov_2017_CVPR} does not explicitly use a multilayer perceptron based encoder-decoder. Conventionally, mean squared error loss is used to reduce the discrepancy between the identical embeddings. However, this may not preserve the semantic structure. In this work, we explicitly formulate objective functions to preserve the semantic structure in the embedding space.

In order to facilitate the task of preserving semantic relations, we consider a tuple of visual features $(\mathbf{x_i},\mathbf{x_j},\mathbf{x_k})$ for every 
class embedding $\mathbf{y_r}$ to be mapped. 
The elements of the tuple are sampled such that the semantic relationship between their corresponding class embeddings and  $\mathbf{y_r}$ satisfy the conditions in the definition. 
Specifically, if $(\mathbf{y_i},\mathbf{y_j}, \mathbf{y_k})$ is the class embedding tuple corresponding to the visual tuple, then $\delta_{ir} = 1$, $\tau \leq \delta_{jr} < 1$ and $\delta_{kr} < \tau$. 
The first feature corresponds to the same class (i.e. $\mathbf{y_r}=\mathbf{y_i}$), the second corresponds to  a semantically similar class and the third feature corresponds to a semantically dissimilar class.
We present a way to efficiently sample the tuples in Section~\ref{mining}. 
Note that this essentially forms a quadruplet $(\mathbf{y_r},\mathbf{x_i},\mathbf{x_j}, \mathbf{x_k})$. 
Though quadruplet based algorithms have been used in literature~\cite{chen2017beyond}, the one explored here is fundamentally different. 
\\\\
\textbf{Objective for Identical and Dissimilar Classes.} The mapped class embedding $f(\mathbf{y_r};\theta_f)$ and the visual feature $\mathbf{x_i}$ must have a high semantic similarity score as they belong to the same class. Ideally, it should be equal to one. 
Also $f(\mathbf{y_r};\theta_f)$ and the visual feature $\mathbf{x_k}$ must have a very low semantic similarity score as they belong to dissimilar classes, i.e. $\delta_{kr} < \tau$. The objective function which caters to the above needs is given by:
\begin{equation}
\label{normal_loss}
\mathcal{O}_1 = \min_{\theta_f}\ \small{ -s\big(f(\mathbf{y_r};\theta_f),\,\mathbf{x_i}\big) + \big(\tau-\delta_{kr} \big)\cdot  s\big(f(\mathbf{y_r};\theta_f),\,\mathbf{x_k}\big)}
\end{equation}   
The first term caters to the identical class and aims to maximize the semantic similarity between $f(\mathbf{y_r}; \theta_f)$ and $\mathbf{x_i}$. 
The second term aims to minimize the semantic similarity between dissimilar entities $f(\mathbf{y_r};\theta_f)$ and $\mathbf{x_k}$. 
Here $(\tau-\delta_{kr})$ acts as an adaptive scaling term, i.e. if the class embeddings are very dissimilar, we put a higher weight on the term to minimize it.\\\\
\textbf{Objective for Similar Classes.} Since $\mathbf{y_r}$ and $\mathbf{y_j}$ are embeddings of semantically similar classes, $f(\mathbf{y_r};\theta_f)$ and $\mathbf{x_j}$ need to be close in order to preserve this relation. 
Explicitly, $s\big(f(\mathbf{y_r}; \theta_f),\,\mathbf{x_j}\big)$ must be greater than $\tau$. In addition to this condition, we also want to ensure that the above enforced condition does not interfere with zero-shot recognition. Therefore, we restrict the semantic similarity score $s\big(f(\mathbf{y_r}; \theta_f),\,\mathbf{x_j}\big)$ to be  less than $\delta_{jr}$. 
This ensures that semantic similarity is preserved without hindering the recognition task. Mathematically, the objective which reflects the above two conditions is as follows:
\begin{equation}
\label{structure}
\mathcal{O}_2 = \min_{\theta_f}\  \small{\big[\tau - s\big(f(\mathbf{y_r}; \theta_f),\,\mathbf{x_j}\big)\big]_+ + \big[s\big(f(\mathbf{y_r};\theta_f),\,\mathbf{x_j}\big) - \delta_{jr}\big]_+}
\end{equation}
where $[z]_+ = \max[0,z]$. Note that only one of the two terms is triggered, corresponding to the either of the conditions $s\big(f(\mathbf{y_r};\theta_f),\,\mathbf{x_j}\big) \geq \tau$ or $s\big(f(\mathbf{y_r};\theta_f),\,\mathbf{x_j}\big) \leq \delta_{jr}$ they violate. 
The above constraints are enforced only on semantically similar classes. 
For semantically dissimilar classes, we aim to have a similarity score as small as possible regardless of the amount of dissimilarity because in most applications the amount of dissimilarity is of little concern.\\\\
\textbf{Reconstruction Loss.}
Since our setup involves a decoder which reconstructs the input $\mathbf{y_r}$, there is an accompanying reconstruction loss. We noted in our experiments that using this additional condition of reconstruction provided better updates to the encoder and enhanced zero-shot recognition performance. In addition, this is in spirit with the observation of Kodirov \etal~\cite{Kodirov_2017_CVPR} that adding an additional reconstruction term is beneficial for zero-shot recognition. 
\begin{equation}
\label{recons}
\mathcal{O}_3 = \min_{\theta_f, \theta_g} ||\mathbf{y_r} - \mathbf{\hat y_r}||_2^2
\end{equation} 
where, $\mathbf{\hat y_r}$ is the output of the decoder $g(\mathbf{x}; \theta_g)$.\\\\
\textbf{Overall Objective.}
With the above three objective functions, the overall objective is given by:
\begin{equation}
\mathcal{O}  = \dfrac{1}{|\mathcal{B}|}\sum_{\mathcal{B}}\mathcal{O}_1 + \lambda_1\mathcal{O}_2 + \lambda_2\mathcal{O}_3 
\end{equation}
Here $|\mathcal{B}|$ refers to the size of the mini-batch $\mathcal{B}$. $\lambda_1$ and $\lambda_2$ are hyper-parameters chosen based on the validation data. 
Given a testing sample $\mathbf{x^u}$, we infer its class as follows:
\begin{equation}
c^* = \arg\max_c \ s\big(f(\mathbf{y_r^c};\theta_f),\, \mathbf{x^u}\big)
\end{equation}
where $\mathbf{y_r^c}$ refers to the class embeddings of only the unseen classes in the conventional zero-shot setting and to the class embeddings of both seen as well as unseen classes in the generalized zero-shot setting.
 \begin{table*}[t!]
 	\centering
 	\footnotesize
 	
 	\begin{tabular}{|c|c|c|c|c|c|c|c|c|c|}
 		\hline
 		\textbf{Dataset} & \begin{tabular}[c]{@{}c@{}}No. of \\ attributes\end{tabular} & \begin{tabular}[c]{@{}c@{}}No. of \\ Seen\\ Classes\end{tabular} & \begin{tabular}[c]{@{}c@{}}No. of \\ unseen\\ classes\end{tabular} & \begin{tabular}[c]{@{}c@{}}No. of \\ samples\end{tabular} & \begin{tabular}[c]{@{}c@{}}No.of \\ samples\\ (Train)\end{tabular} & \begin{tabular}[c]{@{}c@{}}No. of\\ samples from\\ unseen classes\\ (Test)\end{tabular} & \begin{tabular}[c]{@{}c@{}}No. of\\ samples from\\ seen classes\\ (Test)\end{tabular} & \begin{tabular}[c]{@{}c@{}}Hidden layer\\ size (Encoder) \\ \end{tabular} & \begin{tabular}[c]{@{}c@{}}Hidden layer\\ size (Decoder) \\ \end{tabular} \\ \hline
 		\textbf{SUN}~\cite{xiao2010sun}   & 102                                                          & 645                                                              & 72                                                                 & 14340                                                     & 10320                                                              & 1440                                                                                    & 2580                                                                                  & $H_1$ = 1024                                                                     & $H_1$ = 1024                                                                     \\ \hline
 		\textbf{AWA2}~\cite{xian2017zero1}    & 85                                                           & 40                                                               & 10                                                                 & 37322                                                     & 23527                                                              & 7913                                                                                    & 5882                                                                                  & \begin{tabular}[c]{@{}c@{}}$H_1$ = 512\\ $H_2$ = 1024\end{tabular}                  & \begin{tabular}[c]{@{}c@{}}$H_1$ = 1024\\ $H_2$ = 512\end{tabular}                  \\ \hline
 		\textbf{CUB}~\cite{wah2011caltech}     & 312                                                          & 150                                                              & 50                                                                 & 11788                                                     & 7057                                                               & 2967                                                                                    & 1764                                                                                  & $H_1$ = 1024                                                                     & $H_1$ = 1024                                                                     \\ \hline
 		\textbf{aPY}~\cite{farhadi2009describing}     & 64                                                           & 20                                                               & 12                                                                 & 15339                                                     & 5932                                                               & 7924                                                                                    & 1483                                                                                  & \begin{tabular}[c]{@{}c@{}}$H_1$ = 512\\ $H_2$ = 1024\end{tabular}                  & \begin{tabular}[c]{@{}c@{}}$H_1$ = 1024\\ $H_2$ = 512\end{tabular}                  \\ \hline
 	\end{tabular}
 	\caption{\small Details of the datasets with the corresponding encoder-decoder architecture used in the proposed approach. 
 		$H_1$ and $H_2$ refers to the first and second hidden layer respectively. 
 		Only one hidden layer has been used for SUN and CUB datasets.}
 	\label{data-stats}
 \end{table*}
\subsection{Mining the Tuples}
\label{mining}
The proposed algorithm relies on sampling the tuples for preserving the semantic relations. 
In the tuple $(\mathbf{x_i},\mathbf{x_j},\mathbf{x_k})$, $\mathbf{x_i}$ can be chosen at random such that it belongs to the same class as $\mathbf{y_r}$, which we choose sequentially from the dataset.
There are many possible ways to choose $\mathbf{x_j}$ and $\mathbf{x_k}$. 
Choosing the most informative tuples will help in faster convergence and provide useful updates for gradient descent based algorithms. 
In this work, we sample the tuples in an online fashion, wherein for each epoch a criterion is evaluated. Our method is similar to the hard negative mining approach for triplet based learning algorithms \cite{bucher2016hard,schroff2015facenet,simo2015discriminative}.
For every $\mathbf{y_r}$ we wish to embed, we randomly sample $p$ $(p = 50)$ $\mathbf{x_j}'$s which satisfy the condition $\tau\leq\delta_{ij}<1$. Among these $\mathbf{x_j}'$s, we update the parameters of the model with that particular sample which gives the highest loss in objective $\mathcal{O}_2$.
Similarly, we randomly sample $p$ different $\mathbf{x_k}'$s which satisfy the condition that $\delta_{ij}<\tau$. Among these $\mathbf{x_k}'$s, we update the parameters of the model with that particular sample which gives the highest value in the second term of objective $\mathcal{O}_1$. 

It can be seen that we update the model from a set of randomly sampled points. 
This is much more efficient compared to updating the model using the \textit{hardest negative}~\cite{schroff2015facenet} which involves computing the maximum over a much larger set of points. In addition, our method also circumvents the problem of potentially reaching a bad minima due to updation of the model with the \textit{hardest negative}. We also tried updating with the \textit{semi-hard} negatives as described in \cite{schroff2015facenet}, but it did not lead to any significant impact on the results. 

\section{Experiments}
\label{expts}

\subsection{Datasets and Experimental Setting}
We evaluate the proposed approach on four datasets for zero-shot learning : \textbf{SUN}~\cite{xiao2010sun}, Animals with Attributes 2 (\textbf{AWA2})~\cite{xian2017zero1}, Caltech UCSD Birds 200-2011 (\textbf{CUB})~\cite{wah2011caltech} and  Attribute Pascal and Yahoo dataset (\textbf{aPY})~\cite{farhadi2009describing}. All these datasets are provided with annotated attributes. 

The details of these datasets are listed in Table~\ref{data-stats}.
It was observed in \cite{Xian_2017_CVPR} that some of the testing categories in the original split of the datasets are subset of the Imagenet~\cite{deng2009imagenet} categories. Hence, extracting features from Imagenet trained models will not result in a true zero-shot setting. In order to alleviate the problem, the authors proposed a new split such that none of the testing categories coincide with Imagenet categories. In addition, some samples from seen categories were held out for generalized zero-shot recognition. 
Hence, we employ the protocol and the splits as described in \cite{Xian_2017_CVPR,xian2017zero1}.
We use continuous per-class attributes for all the datasets and average per-class top-1 accuracy to report the results.

We use the 2048-D Resnet-101~\cite{he2016deep} features provided by \cite{Xian_2017_CVPR} for all the datasets. 
The architecture details of the proposed approach are given in Table \ref{data-stats}. 
ReLU activation is used for all the layers except for the output of the encoder and the decoder, which employ ELU~\cite{clevert2015fast} activations. We use Adam optimizer~\cite{kingma2014adam} with a learning rate of $10^{-3}$ and a weight decay of $5\times10^{-5}$. 
All the input features and attributes are normalized to have zero mean and unit standard deviation.

The comparisons with the state-of-the-art are made with the results reported in \cite{xian2017zero1}, as it is based on exactly the same protocol and use the same set of features. Besides, the algorithms on which the results are reported encompass a wide range of approaches in zero-shot learning. \\ \\
{\bf Baselines.} We define three baseline settings which provide insights into the importance of each of the terms in the proposed objective function. 
For baseline $\mathbf{B_1}$, instead of  $\mathcal{O}_1$ and $\mathcal{O}_2$, mean squared error objective is used to learn the mapping.
The same architecture as used by the proposed approach is used along with the reconstruction objective $\mathcal{O}_3$.  
This will align the mapped class embeddings with the structure of the visual space.
Since this setup does not enforce the relations as described in Section~\ref{relations}, the embedding space may not maintain the structure of the semantic space.
For baseline $\mathbf{B_2}$, we set $\tau = 1$ and $\lambda_1 = 0$ in our approach. This helps us to better understand the importance of objective $\mathcal{O}_2$.
This results in maximizing the cosine similarity for identical classes and minimizing the same for all other classes.
In addition to the above two baselines, we also demonstrate the importance of the objective $\mathcal{O}_3$. We employ just the encoder and set $\lambda_2=0$ in our experiments for $\mathbf{B_3}$. 
This provides insight into the degree of enhancement in performance due to the reconstruction term.

\subsection{Conventional Zero-Shot Learning Results}
The results of the proposed approach on various datasets is listed in Table~\ref{zsl-results}. 
We also provide comparisons with the state-of-the-art. 
With respect to the first baseline B\textsubscript{1}, we observe that the proposed approach consistently performs better on all the datasets. 
This supports our hypothesis that inheriting semantic properties to the embedding space is beneficial for zero-shot recognition.  
We also observe significant increase in performance when we include the objective $\mathcal{O}_2$ in our approach.
In this case, the difference in performance is pronounced in coarse-grained datasets AWA2 and aPY wherein the inter-class semantics are much different. 
This indicates that the objective $\mathcal{O}_2$, which is essentially the structure preserving term, is beneficial for zero-shot recognition. 
The reconstruction term also contributes to varying levels of gain in performance. Visualization of the embedding space is presented in Figure \ref{tsne-awa} for the ten unseen classes of AWA2 dataset. It can be seen that the semantic relations are preserved to a good extent. 

\begin{figure}[t!]
	\centering
	\includegraphics[width=8cm, height=5cm]{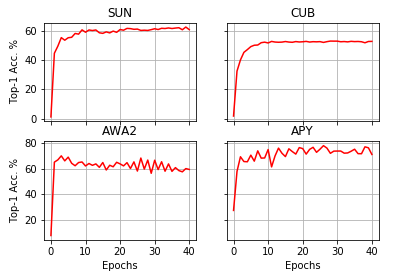}
	\caption{\small Plot of validation accuracy against number of epochs.}
	\label{val-plot} 
\end{figure}

The proposed approach also compares favorably with the existing approaches in literature, with our approach obtaining the state-of-the-art on SUN, AWA2 and CUB datasets. 
On aPY dataset, we obtain 38.4\% which is slightly less than Deep Visual Semantic Embedding Model~\cite{frome2013devise}. 
However, on the generalized zero-shot setting, the proposed approach performs much better, as illustrated next.\\\\
\textbf{Effectiveness of the tuple mining approach.} The graph showing the accuracy on the validation split against the number of epochs for the four datasets is shown in Figure~\ref{val-plot}. 
We observe that for all the datasets, around 80\% of the maximum accuracy is reached in less than 5 epochs. 
 
\begin{table}[t!]
	\centering
	\small
	\begin{tabular}{|c|c|c|c|c|}
		\hline
		\textbf{Method}            & \textbf{SUN}  & \textbf{AWA2} & \textbf{CUB}  & \textbf{aPY}  \\ \hline
		DAP~\cite{lampert2014attribute}               & 39.9          & 46.1          & 40.0          & 33.8          \\ 
		IAP~\cite{lampert2014attribute}              & 19.4          & 35.9          & 24.0          & 36.6          \\ 
		CONSE~\cite{norouzi2013zero}             & 38.8          & 44.5          & 34.3          & 26.9          \\ 
		CMT~\cite{socher2013zero}               & 39.9          & 37.9          & 34.6          & 28.0          \\ 
		SSE~\cite{zhang2015zero}               & 51.5          & 61.0          & 43.9          & 34.0          \\ 
		LATEM~\cite{xian2016latent}             & 55.3          & 55.8          & 49.3          & 35.2          \\ 
		ALE~\cite{akata2016label}               & 58.1          & 62.5          & 54.9          & 39.7          \\ 
		DEVISE~\cite{frome2013devise}            & 56.5          & 59.7          & 52.0          & \textbf{39.8} \\ 
		SJE~\cite{akata2015evaluation}               & 53.7          & 61.9          & 53.9          & 32.9          \\ 
		ESZSL~\cite{romera2015embarrassingly}             & 54.5          & 58.6          & 53.9          & 38.3          \\ 
		SYNC~\cite{changpinyo2016synthesized}              & 56.3          & 46.6          & 55.6          & 23.9          \\ 
		SAE~\cite{Kodirov_2017_CVPR}               & 40.3          & 54.1          & 33.3          & 8.3           \\ \hline
		MSE + Recons. ($\mathbf{B_1}$)&58.5   &54.9    &49.2  &34.8 \\ 
		Proposed - $\mathcal{O}_2$ ($\mathbf{B_2}$)  &57.1 &57.2 &51.5 &31.6\\ 
		Proposed - $\mathcal{O}_3$ ($\mathbf{B_3}$)  &58.7 &62.4 &52.7 &37.2\\ \hline
		\textbf{Proposed} & \textbf{61.4} & \textbf{63.8} & \textbf{56.0} & 38.4          \\ \hline 
	\end{tabular}
	\caption{\small Average per-class accuracy (top-1 in \%) for conventional zero-shot learning. Results of the existing approaches are taken from \cite{xian2017zero1}.}
	\label{zsl-results}
\end{table}
\subsection{Generalized Zero-Shot Learning Results}
\begin{table*}[t!]
	\centering
	\small
	\begin{tabular}{|c|c|c|c|c|c|c|c|c|c|c|c|c|}
		\hline
		& \multicolumn{3}{c|}{\textbf{SUN}}             & \multicolumn{3}{c|}{\textbf{AWA2}}            & \multicolumn{3}{c|}{\textbf{CUB}}             & \multicolumn{3}{c|}{\textbf{aPY}}    \\ \hline
		\textbf{Method}   & \textbf{ts}            & \textbf{tr}            & \textbf{H}             & \textbf{ts}            & \textbf{tr}            & \textbf{H}             & \textbf{ts}            & \textbf{tr}            & \textbf{H}             & \textbf{ts}            & \textbf{tr}   & \textbf{H}             \\ \hline
		DAP~\cite{lampert2014attribute}                & 4.2           & 25.1          & 7.2           & 0.0           & 84.7          & 0.0           & 1.7           & 67.9          & 3.3           & 4.8           & 78.3 & 9.0           \\
		IAP~\cite{lampert2014attribute}                & 1.0           & 37.8          & 1.8           & 0.9           & 87.6          & 1.8           & 0.2           & \textbf{72.8} & 0.4           & 5.7           & 65.6 & 10.4          \\
		CONSE~\cite{norouzi2013zero}             & 6.8           & 39.9 & 11.6          & 0.5           & \textbf{90.6} & 1.0           & 1.6           & 72.2          & 3.1           & 0.0           & \textbf{91.2} & 0.0           \\
		CMT~\cite{socher2013zero}               & 8.1           & 21.8          & 11.8          & 0.5           & 90.0          & 1.0           & 7.2           & 49.8          & 12.6          & 1.4           & 85.2 & 2.8           \\
		CMT*~\cite{socher2013zero}              & 8.7           & 28.0          & 13.3          & 8.7           & 89.0          & 15.9          & 4.7           & 60.1          & 8.7           & 10.9          & 74.2 & 19.0          \\
		SSE~\cite{zhang2015zero}               & 2.1           & 36.4          & 4.0           & 8.1           & 82.5          & 14.8          & 8.5           & 46.9          & 14.4          & 0.2           & 78.9 & 0.4           \\
		LATEM~\cite{xian2016latent}             & 14.7          & 28.8          & 19.5          & 11.5          & 77.3          & 20.0          & 15.2          & 57.3          & 24.0          & 0.1           & 73.0 & 0.2           \\
		ALE~\cite{akata2016label}                & \textbf{21.8} & 33.1          & 26.3          & 14.0          & 81.8          & 23.9          & 23.7          & 62.8          & \textbf{34.4} & 4.6           & 73.7 & 8.7           \\
		DEVISE~\cite{frome2013devise}            & 16.9          & 27.4          & 20.9          & 17.1          & 74.7          & 27.8          & 23.8          & 53.0          & 32.8          & 4.9           & 76.9 & 9.2           \\
		SJE~\cite{akata2015evaluation}               & 14.7          & 30.5          & 19.8          & 8.0           & 73.9          & 14.4          & 23.5          & 59.2          & 33.6          & 3.7           & 55.7 & 6.9           \\
		ESZSL~\cite{romera2015embarrassingly}             & 11.0          & 27.9          & 15.8          & 5.9           & 77.8          & 11.0          & 12.6          & 63.8          & 21.0          & 2.4           & 70.1 & 4.6           \\
		SYNC~\cite{changpinyo2016synthesized}              & 7.9           & \textbf{43.3}          & 13.4          & 10.0          & 90.5          & 18.0          & 11.5          & 70.9          & 19.8          & 7.4           & 66.3 & 13.3          \\
		SAE~\cite{Kodirov_2017_CVPR}               & 8.8           & 18.0          & 11.8          & 1.1           & 82.2          & 2.2           & 7.8           & 54.0          & 13.6          & 0.4           & 80.9 & 0.9           \\ \hline
		MSE + Recons. ($\mathbf{B_1}$)& 12.8          & 38.9          &  19.3 & 13.6 & 72.4          & 22.9 & 13.1 & 48.8          & 20.7        & 10.9 & 51.3 & 18.0 \\
		Proposed - $\mathcal{O}_2$ ($\mathbf{B_2}$)&17.2 &35.3 &23.1 &15.3  &73.5 &25.3 &20.7 &51.6 &29.5 &7.6 &36.0 &12.6\\ 
		Proposed - $\mathcal{O}_3$ ($\mathbf{B_3}$)&16.9 &34.2 &22.4 &20.5  &72.9 &32.0 &20.9 &52.3 &29.9 &11.4 &48.7 &18.5\\ \hline
		\textbf{Proposed} & 20.8          & 37.2          & \textbf{26.7} & \textbf{20.7} & 73.8          & \textbf{32.3} & \textbf{24.6} & 54.3          & 33.9          & \textbf{13.5} & 51.4 & \textbf{21.4} \\ \hline
	\end{tabular}
	\caption{\small Results on generalized zero-shot learning. \textbf{ts} refers to the setting wherein the testing samples belong to unseen classes. \textbf{tr} refers to the setting in which the testing samples belong to either seen classes or unseen classes. \textbf{H} refers to the harmonic mean between \textbf{ts} and \textbf{tr}. The results of the existing approaches are taken from \cite{xian2017zero1}. CMT* refers to CMT~\cite{socher2013zero} with novelty detection.}
	\label{gen-zsl-results}
\end{table*}
In \cite{Xian_2017_CVPR}, some of the samples from seen classes are held out for testing. In this setting, the search space consists of both the seen classes as well as the unseen classes. This scenario is more realistic, as we cannot usually anticipate whether an incoming sample belongs to a seen class or an unseen class. Table~\ref{gen-zsl-results} reports the result of generalized zero-shot learning on the four datasets under two different settings. 
The first setting (referred to as \textbf{ts}) involves comparison of samples from unseen classes against both seen and unseen classes. 
The second setting (referred to as \textbf{tr}) involves comparison of samples from unseen classes as well as held-out samples from seen classes against all the classes. High accuracy on \textbf{tr} and low accuracy on \textbf{ts} implies that the model performs well on the seen classes but fails to generalize to the unseen classes. 
The harmonic mean (denoted by \textbf{H}) of the two results is also reported, as this measure encourages accuracies for both the settings to be high~\cite{Xian_2017_CVPR,xian2017zero1}.

With respect to B\textsubscript{1}, we can see that our method performs better in the first setting (\textbf{ts}) by a large margin. With respect to B\textsubscript{2}, there is a gain in accuracy on all the settings. The difference is pronounced in the first setting, which is concerned with samples belonging to novel classes. This indicates that employing the concept of semantic relations and preserving these relations in the embedding space is beneficial for classification on novel categories. In fact, the observations made for conventional zero-shot learning setting are also applicable here in a more realistic setting. With respect to the state-of-the-art, our approach gives a harmonic mean accuracy of 26.7\%  on SUN which is the best result among all the reported methods. 
In addition, we obtain 32.3\% on the AWA2 dataset, better than the next best method by nearly 5\%. On CUB, proposed approach obtains a best accuracy of 24.6\% on the first setting and 33.9\% overall. On aPY dataset, our approach achieves 13.5\% on the first setting and 51.4\% on the second setting, with an overall result of 21.4\%. It can be seen that methods like CONSE~\cite{norouzi2013zero} perform very well on seen classes but do not generalize well for novel classes. Although our method does not match the accuracy of the methods like CONSE~\cite{norouzi2013zero} and CMT~\cite{socher2013zero} on the second setting, it outperforms them on the first setting by a large margin which is reflected in the harmonic mean of the two. In addition, our method performs better compared to other competitive methods like ALE~\cite{akata2016label} and DEVISE~\cite{frome2013devise} on the first setting on AWA2, CUB and aPY. 
\subsection{Experiments on ImageNet}
\label{imnet}
\begin{table}[]
	\centering
	\small
	\begin{tabular}{|c|c|c|c|c|c|c|c|}
		\hline
		\multicolumn{1}{|l|}{\multirow{2}{*}{}} & \multicolumn{1}{l|}{\multirow{2}{*}{}} & \multicolumn{3}{c|}{\begin{tabular}[c]{@{}c@{}}\textbf{Conventional} \\ \textbf{ZSL}\end{tabular}} & \multicolumn{3}{c|}{\begin{tabular}[c]{@{}c@{}}\textbf{Generalized}\\  \textbf{ZSL}\end{tabular}} \\ \cline{3-8} 
		\multicolumn{1}{|l|}{}                  & \multicolumn{1}{l|}{}                  & \textbf{2H}                      & \textbf{3H}                     & \textbf{All}                     & \textbf{2H}                     & \textbf{3H}                     & \textbf{All }                    \\ \hline
		\multirow{2}{*}{Top-1}                  & SYNC~\cite{changpinyo2016synthesized}                                   &  9.1                          &  2.6                         &  0.9                       &    0.3                       &  0.1                         &     0.0                    \\ \cline{2-8} 
		& \textbf{Proposed}                               &    9.4                        & 2.8                          &  1.0                      &  1.2                         &          0.8                 &        0.4                 \\ \hline
		\multirow{2}{*}{Top-5}                  & SYNC~\cite{changpinyo2016synthesized}                                   &  25.9                          &  4.9                         &    2.5                   &  8.7                         &   3.8                        &     1.2                    \\ \cline{2-8} 
		& \textbf{Proposed}                               &  26.3                          & 4.8                          &2.7                         & 11.2                         &    4.9                       &       1.7                  \\ \hline
	\end{tabular}
	\caption{\small Results on ImageNet. We measure top-1 and top-5 accuracies  in \%. 2H / 3H refers to the test split in which classes are 2 / 3 tree hops away from train classes in the WordNet hierarchy.}
	\label{imnet-results}
\end{table}
\begin{figure*}[]
	\centering
	\includegraphics[width=17.5cm, height=7.5cm]{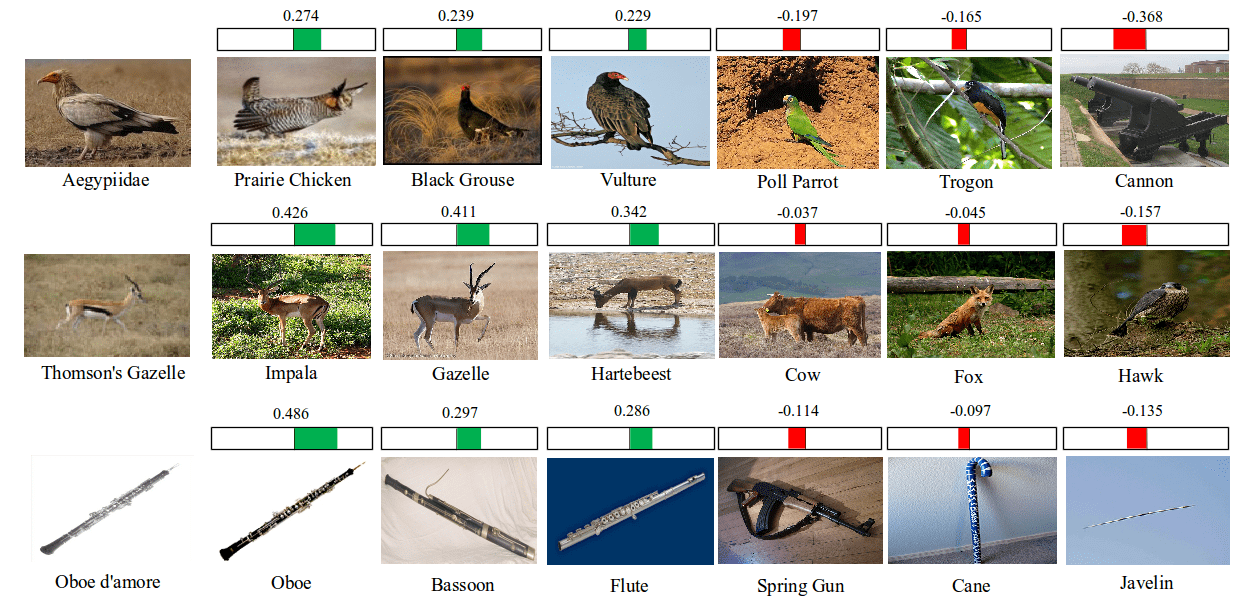}
	\caption{\small Results on approximate semantic inference. The image at the left corresponds to the query image for which class embeddings are not available. Comparison using cosine similarity with the existing class embeddings (seen as well as unseen) is shown. The numerical figures indicate the cosine similarity of the image with respect to the class. The bar indicates whether the model predicts it as similar (green) or dissimilar (red). $\tau=0$  for this setup.}
	\label{cosine_sim} 
\end{figure*}
ImageNet~\cite{deng2009imagenet} is a large scale dataset consisting of nearly 14.1 million images belonging to 21,841 categories. The 1000 categories of ILSRVC are used as seen classes\footnote{Images from one of the seen classes namely `teddy bear' which belongs to the synset n04399382 was unavailable. Hence, we use only 999 categories for training.} while the rest as unseen classes. There are no curated attributes available for this dataset. 
Distributed word embeddings like Word2Vec~\cite{mikolov2013distributed} are employed instead as they have been shown to contain semantic properties which are suitable for zero-shot learning. We extract the Resnet-101~\cite{he2016deep} features from the pretrained model available in the pytorch~\cite{pytorch} model zoo. We use the Word2Vec provided by~\cite{changpinyo2016synthesized}. We employ two hidden layers for the encoder and the decoder, similar to AWA2 and aPY datasets. 
For comparison, we implement the SYNC~\cite{changpinyo2016synthesized} algorithm with the aforementioned features and settings. To the best of our knowledge, SYNC achieves the state-of-the-art performance on this very challenging dataset~\cite{Xian_2017_CVPR,xian2017zero1}. We use the public code made available by the authors for evaluation.

Table \ref{imnet-results} lists the results obtained using the proposed approach on different splits of test data. 
We observe that our approach consistently achieves better performance compared to SYNC, thus improving over the state-of-the-art.
On the generalized zero-shot setting, we observe that the proposed approach outperforms SYNC by a large margin. This indicates that the proposed approach scales favorably to a realistic setting with large number of unseen classes. 
\subsection{Approximate Semantic Inference}
\label{approx_semantic_inference} 
It is reasonable to assume that attributes or other form of side information is available in most of the scenarios. However, there might be a situation in which side information for a few categories of interest might not be available. For example, in ImageNet, out of the 21,841 categories, Word2Vec for 497 categories is not present. This bottleneck might crop up in a very large scale setting as evident in the ImageNet example. In this scenario, though classification is not possible, we wish to infer the approximate semantic characteristics of the object in the image. Since our model has semantic characteristics in its embedding space, we can approximately infer the semantic properties of the image in \textit{relation} to the existing categories (seen and unseen). Some of the empirical results on categories of ImageNet for which class embeddings is not available can be seen in Figure \ref{cosine_sim}. 
In the first example, the bird in the image belongs to \textit{Aegypiidae}, for which the class embedding is not available. The image feature is compared using cosine similarity with the existing class embeddings. It can be inferred that the given image is \textit{semantically similar} to \textit{Prairie Chicken}, \textit{Black Chicken} and \textit{Vulture}. Moreover, it can also be inferred that categories \textit{Poll Parrot}, \textit{Trogon} and \textit{Cannon} are dissimilar to the bird in the  image. This is very much in agreement to the actual semantic properties of the categories. In addition, the cosine similarity score is evocative of the degree of similarity between the actual category and the category with which it is compared. Similar observations can be made from other examples as well. This suggests that inheriting the structure of semantic space helps to make approximate inference about  an image with respect to known entities, thus showing potential for tasks beyond just zero-shot classification.
\section{Discussion}
\label{discuss}
The cosine similarity function applied on the mapped class embeddings and the image features can be approximated to a normalized compatibility score function.  Thus, the setup of baseline B\textsubscript{2} is similar to the ranking based methods~\cite{akata2013label,xian2016latent,frome2013devise} which employ compatibility functions wherein the embeddings which belong to the same class are pulled together while the rest are pushed apart. 
The results are also similar to the ones achieved using these methods. Thus, a particular instantiation of the proposed approach can be approximated to compatibility models with the ranking objective.
In addition, the merits of approaches \cite{Kodirov_2017_CVPR} and \cite{zhang2016learning} are seamlessly incorporated in our model.

Although the advantages of the proposed approach is clear and encouraging, one of the limitations is its performance on the seen categories in the generalized zero-shot setting. Though it does not match the results on some of the previous approaches in literature, the proposed approach still gives encouraging performance. We believe that exploration of more intricate forms of relations between categories would help in furthering the state-of-the art in this setting.

\section{Conclusion}
\label{conclude}
In this work, we focus on efficiently utilizing the structure of the semantic space for improved classification on the unseen categories. 
We introduce the concept of relations between classes in terms of their semantic content. 
We devise objective functions which help in preserving semantic relations in the embedding space thereby inheriting the structure of the semantic space.
Extensive evaluation of the proposed approach is carried out and state-of-the-art results are obtained on multiple settings including the tougher generalized zero-shot learning, thus proving its effectiveness for zero-shot learning.\\\\
\textbf{Acknowledgment.} The authors would like to thank Devraj Mandal of IISc for helpful discussions.
{\small
	\bibliographystyle{ieee}
	\bibliography{egbib}

\begin{thebibliography}{10}\itemsep=-1pt

\bibitem{numObj}
\url{https://www.nytimes.com/2014/05/27/science/welcoming-the-newly-discovered.html}.

\bibitem{pytorch}
\url{https://github.com/pytorch/pytorch}.

\bibitem{akata2013label}
Z.~Akata, F.~Perronnin, Z.~Harchaoui, and C.~Schmid.
\newblock Label-embedding for attribute-based classification.
\newblock In {\em CVPR}, 2013.

\bibitem{akata2016label}
Z.~Akata, F.~Perronnin, Z.~Harchaoui, and C.~Schmid.
\newblock Label-embedding for image classification.
\newblock {\em PAMI}, 2016.

\bibitem{akata2015evaluation}
Z.~Akata, S.~Reed, D.~Walter, H.~Lee, and B.~Schiele.
\newblock Evaluation of output embeddings for fine-grained image
  classification.
\newblock In {\em CVPR}, 2015.

\bibitem{bucher2016hard}
M.~Bucher, S.~Herbin, and F.~Jurie.
\newblock Hard negative mining for metric learning based zero-shot
  classification.
\newblock In {\em ECCV Workshops}, 2016.

\bibitem{changpinyo2016synthesized}
S.~Changpinyo, W.-L. Chao, B.~Gong, and F.~Sha.
\newblock Synthesized classifiers for zero-shot learning.
\newblock In {\em CVPR}, 2016.

\bibitem{chen2017beyond}
W.~Chen, X.~Chen, J.~Zhang, and K.~Huang.
\newblock Beyond triplet loss: a deep quadruplet network for person
  re-identification.
\newblock {\em CVPR}, 2017.

\bibitem{clevert2015fast}
D.-A. Clevert, T.~Unterthiner, and S.~Hochreiter.
\newblock Fast and accurate deep network learning by exponential linear units
  (elus).
\newblock {\em arXiv preprint arXiv:1511.07289}, 2015.

\bibitem{farhadi2010attribute}
A.~Farhadi, I.~Endres, and D.~Hoiem.
\newblock Attribute-centric recognition for cross-category generalization.
\newblock In {\em CVPR}, 2010.

\bibitem{farhadi2009describing}
A.~Farhadi, I.~Endres, D.~Hoiem, and D.~Forsyth.
\newblock Describing objects by their attributes.
\newblock In {\em CVPR}, 2009.

\bibitem{ferrari2008learning}
V.~Ferrari and A.~Zisserman.
\newblock Learning visual attributes.
\newblock In {\em NIPS}, 2008.

\bibitem{frome2013devise}
A.~Frome, G.~S. Corrado, J.~Shlens, S.~Bengio, J.~Dean, T.~Mikolov, et~al.
\newblock Devise: A deep visual-semantic embedding model.
\newblock In {\em NIPS}, 2013.

\bibitem{he2016deep}
K.~He, X.~Zhang, S.~Ren, and J.~Sun.
\newblock Deep residual learning for image recognition.
\newblock In {\em CVPR}, 2016.

\bibitem{kingma2014adam}
D.~Kingma and J.~Ba.
\newblock Adam: A method for stochastic optimization.
\newblock {\em arXiv preprint arXiv:1412.6980}, 2014.

\bibitem{Kodirov_2017_CVPR}
E.~Kodirov, T.~Xiang, and S.~Gong.
\newblock Semantic autoencoder for zero-shot learning.
\newblock In {\em CVPR}, 2017.

\bibitem{lampert2009learning}
C.~H. Lampert, H.~Nickisch, and S.~Harmeling.
\newblock Learning to detect unseen object classes by between-class attribute
  transfer.
\newblock In {\em CVPR}, 2009.

\bibitem{lampert2014attribute}
C.~H. Lampert, H.~Nickisch, and S.~Harmeling.
\newblock Attribute-based classification for zero-shot visual object
  categorization.
\newblock {\em PAMI}, 2014.

\bibitem{lei2015predicting}
J.~Lei~Ba, K.~Swersky, S.~Fidler, and R.~Salakhutdinov.
\newblock Predicting deep zero-shot convolutional neural networks using textual
  descriptions.
\newblock In {\em ICCV}, 2015.

\bibitem{maaten2008visualizing}
L.~v.~d. Maaten and G.~Hinton.
\newblock Visualizing data using t-sne.
\newblock {\em JMLR}, 2008.

\bibitem{mikolov2013efficient}
T.~Mikolov, K.~Chen, G.~Corrado, and J.~Dean.
\newblock Efficient estimation of word representations in vector space.
\newblock {\em arXiv preprint arXiv:1301.3781}, 2013.

\bibitem{mikolov2013distributed}
T.~Mikolov, I.~Sutskever, K.~Chen, G.~S. Corrado, and J.~Dean.
\newblock Distributed representations of words and phrases and their
  compositionality.
\newblock In {\em NIPS}, 2013.

\bibitem{Morgado_2017_CVPR}
P.~Morgado and N.~Vasconcelos.
\newblock Semantically consistent regularization for zero-shot recognition.
\newblock In {\em CVPR}, 2017.

\bibitem{norouzi2013zero}
M.~Norouzi, T.~Mikolov, S.~Bengio, Y.~Singer, J.~Shlens, A.~Frome, G.~S.
  Corrado, and J.~Dean.
\newblock Zero-shot learning by convex combination of semantic embeddings.
\newblock {\em arXiv preprint arXiv:1312.5650}, 2013.

\bibitem{palatucci2009zero}
M.~Palatucci, D.~Pomerleau, G.~E. Hinton, and T.~M. Mitchell.
\newblock Zero-shot learning with semantic output codes.
\newblock In {\em NIPS}, 2009.

\bibitem{parikh2011relative}
D.~Parikh and K.~Grauman.
\newblock Relative attributes.
\newblock In {\em ICCV}, 2011.

\bibitem{pennington2014glove}
J.~Pennington, R.~Socher, and C.~D. Manning.
\newblock Glove: Global vectors for word representation.
\newblock In {\em EMNLP}, 2014.

\bibitem{qiao2016less}
R.~Qiao, L.~Liu, C.~Shen, and A.~van~den Hengel.
\newblock Less is more: zero-shot learning from online textual documents with
  noise suppression.
\newblock In {\em CVPR}, 2016.

\bibitem{radovanovic2010hubs}
M.~Radovanovi{\'c}, A.~Nanopoulos, and M.~Ivanovi{\'c}.
\newblock Hubs in space: Popular nearest neighbors in high-dimensional data.
\newblock {\em JMLR}, 2010.

\bibitem{radovanovic2010existence}
M.~Radovanovi{\'c}, A.~Nanopoulos, and M.~Ivanovi{\'c}.
\newblock On the existence of obstinate results in vector space models.
\newblock In {\em International ACM SIGIR conference on Research and
  development in information retrieval}, 2010.

\bibitem{romera2015embarrassingly}
B.~Romera-Paredes and P.~Torr.
\newblock An embarrassingly simple approach to zero-shot learning.
\newblock In {\em ICML}, 2015.

\bibitem{deng2009imagenet}
O.~Russakovsky, J.~Deng, H.~Su, J.~Krause, S.~Satheesh, S.~Ma, Z.~Huang,
  A.~Karpathy, A.~Khosla, M.~Bernstein, A.~C. Berg, and L.~Fei-Fei.
\newblock {ImageNet Large Scale Visual Recognition Challenge}.
\newblock {\em IJCV}, 2015.

\bibitem{schroff2015facenet}
F.~Schroff, D.~Kalenichenko, and J.~Philbin.
\newblock Facenet: A unified embedding for face recognition and clustering.
\newblock In {\em CVPR}, 2015.

\bibitem{shigeto2015ridge}
Y.~Shigeto, I.~Suzuki, K.~Hara, M.~Shimbo, and Y.~Matsumoto.
\newblock Ridge regression, hubness, and zero-shot learning.
\newblock In {\em Joint European Conference on Machine Learning and Knowledge
  Discovery in Databases}, 2015.

\bibitem{simo2015discriminative}
E.~Simo-Serra, E.~Trulls, L.~Ferraz, I.~Kokkinos, P.~Fua, and F.~Moreno-Noguer.
\newblock Discriminative learning of deep convolutional feature point
  descriptors.
\newblock In {\em ICCV}, 2015.

\bibitem{socher2013zero}
R.~Socher, M.~Ganjoo, C.~D. Manning, and A.~Ng.
\newblock Zero-shot learning through cross-modal transfer.
\newblock In {\em NIPS}, 2013.

\bibitem{tomasev2014role}
N.~Tomasev, M.~Radovanovic, D.~Mladenic, and M.~Ivanovic.
\newblock The role of hubness in clustering high-dimensional data.
\newblock {\em KDE}, 2014.

\bibitem{wah2011caltech}
C.~Wah, S.~Branson, P.~Welinder, P.~Perona, and S.~Belongie.
\newblock The caltech-ucsd birds-200-2011 dataset.
\newblock 2011.

\bibitem{xian2016latent}
Y.~Xian, Z.~Akata, G.~Sharma, Q.~Nguyen, M.~Hein, and B.~Schiele.
\newblock Latent embeddings for zero-shot classification.
\newblock In {\em CVPR}, 2016.

\bibitem{xian2017zero1}
Y.~Xian, C.~H. Lampert, B.~Schiele, and Z.~Akata.
\newblock Zero-shot learning-a comprehensive evaluation of the good, the bad
  and the ugly.
\newblock {\em arXiv preprint arXiv:1707.00600}, 2017.

\bibitem{Xian_2017_CVPR}
Y.~Xian, B.~Schiele, and Z.~Akata.
\newblock Zero-shot learning - the good, the bad and the ugly.
\newblock In {\em CVPR}, 2017.

\bibitem{xiao2010sun}
J.~Xiao, J.~Hays, K.~A. Ehinger, A.~Oliva, and A.~Torralba.
\newblock Sun database: Large-scale scene recognition from abbey to zoo.
\newblock In {\em CVPR}, 2010.

\bibitem{Xu_2017_CVPR}
X.~Xu, F.~Shen, Y.~Yang, D.~Zhang, H.~Tao~Shen, and J.~Song.
\newblock Matrix tri-factorization with manifold regularizations for zero-shot
  learning.
\newblock In {\em CVPR}, 2017.

\bibitem{zhang2016learning}
L.~Zhang, T.~Xiang, and S.~Gong.
\newblock Learning a deep embedding model for zero-shot learning.
\newblock {\em CVPR}, 2017.

\bibitem{zhang2015zero}
Z.~Zhang and V.~Saligrama.
\newblock Zero-shot learning via semantic similarity embedding.
\newblock In {\em ICCV}, 2015.

\bibitem{zhang2016zero}
Z.~Zhang and V.~Saligrama.
\newblock Zero-shot learning via joint latent similarity embedding.
\newblock In {\em CVPR}, 2016.

\end{thebibliography}
}
\end{document}